\documentclass{article} 
\usepackage{iclr2017_conference,times}
\usepackage{hyperref}
\usepackage{url}
\usepackage{graphicx}

\title{Deep learning analysis of breast MRIs for prediction of occult invasive disease in ductal carcinoma in situ}

\author{Zhe Zhu, Michael Harowicz, Jun Zhang, Ashirbani Saha, Lars J. Grimm\\
Department of Radiology \\
Duke University \\
Durham, NC 27705, USA \\
\texttt{zhe.zhu@duke.edu, michael.harowicz@gmail.com} \\
\texttt{\{jun.zhang.2017,as698,lars.grimm\}@duke.edu} \\
\And
E.Shelley Hwang  \\
Department of Surgery \\
Duke University Comprehensive Cancer Center \\
Durham, NC 27705, USA \\
\texttt{shelley.hwang@duke.edu}
\And
Maciej A. Mazurowski \\
Department of Radiology \\
Duke University \\
Durham, NC 27705, USA \\
\texttt{maciej.mazurowski@duke.edu}
}

\begin{document}

\maketitle

\begin{abstract}
  \textbf{Purpose}: To determine whether deep learning-based algorithms applied to breast MR images can aid in the prediction of occult invasive disease following the diagnosis of ductal carcinoma in situ (DCIS) by core needle biopsy.

\textbf{Material and Methods}: In this institutional review board-approved study, we analyzed dynamic contrast-enhanced fat-saturated T1-weighted MRI sequences of 131 patients at our institution with a core needle biopsy-confirmed diagnosis of DCIS. The patients had no preoperative therapy before breast MRI and no prior history of breast cancer. We explored two different deep learning approaches to predict whether there was a hidden (occult) invasive component in the analyzed tumors that was ultimately detected at surgical excision. In the first approach, we adopted the transfer learning strategy, in which a network pre-trained on a large dataset of natural images is fine-tuned with our DCIS images. Specifically, we used the GoogleNet model pre-trained on the ImageNet dataset. In the second approach, we used a pre-trained network to extract deep features, and a support vector machine (SVM) that utilizes these features to predict the upstaging of the DCIS. We used 10-fold cross validation and the area under the ROC curve (AUC) to estimate the performance of the predictive models.

\textbf{Results}: The best classification performance was obtained using the deep features approach with GoogleNet model pre-trained on ImageNet as the feature extractor and a polynomial kernel SVM used as the classifier (AUC = 0.70, 95\% CI: 0.58-0.79). For the transfer learning based approach, the highest AUC obtained was 0.53 (95\% CI: 0.41-0.62).

\textbf{Conclusion}: Convolutional neural networks could potentially be used to identify occult invasive disease in patients diagnosed with DCIS at the initial core needle biopsy.

\end{abstract}

\section{Introduction}
While most breast cancers are diagnosed via a core needle biopsy (CNB), the limited size, number, and location of samples can lead to undersampling which can miss occult invasive disease. For ductal carcinoma in situ (DCIS), upstaging to invasive disease at surgical excision occurs in roughly one quarter of cases~\cite{Brennan2011}. Failure to diagnose invasive disease prior to surgery can have multiple clinical implications. By definition, DCIS has no metastatic potential so evaluation of regional lymph nodes or distant sites of disease are not performed initially~\cite{VanRoozendaal2016}. Treatment options are different between DCIS and invasive disease, so patients may need to undergo additional surgical procedures if invasive disease is missed. Clinical outcomes are worse for invasive disease, so conversations with patients must subsequently change~\cite{Martinez-Perez2017}.  Lastly, some patients with DCIS may elect to forgo surgery and pursue active surveillance, but the presence of occult invasive disease makes this management strategy riskier~\cite{Grimm2017}.

While multiple efforts have been aimed at using pathology and mammographic features to predict upstaging~\cite{Brennan2011}, there has been limited work utilizing MRI which boasts the greatest sensitivity for cancer detection. Harowicz et al has shown that hand-crafted MRI features~\cite{Harowicz2017} could be effective in predicting DCIS upstaging. However, deep learning offers a promising alternative means of identifying occult invasive disease, based on promising results for translating natural image classification~\cite{Krizhevsky2012,Szegedy2015} to medical image classification~\cite{Shin2016}.

The success of deep learning approaches in natural image processing is primarily the result of diverse and sufficient training samples. For example in the ImageNet~\cite{JiaDeng2009} classification challenge, there are up to 1000 categories of images with 1000 examples for each category, resulting in 1 million training images in total. However, medical imaging datasets are much smaller which hinders the training of convolutional neural networks, due to over-fitting. Nevertheless, deep learning-based methods have been applied to medical imaging for segmentation~\cite{Gao2016}, detection~\cite{Navab2015}, and classification~\cite{Gao2016a} tasks by utilizing problem-specific modifications. For example, to predict the pixels in the border region in biomedical image segmentation, the missing context is extrapolated by mirroring the input image~\cite{Ronneberger2015}. For more details reader can refer to the surveys~\cite{Litjens2017,Greenspan2016} of deep learning in medical imaging.

In this study, we applied deep learning techniques to breast MRIs to identify which patients diagnosed with DCIS at core needle biopsy have occult invasive disease. For this purpose, we evaluated two approaches. In the transfer learning approach, instead of random initialization, a pre-trained model was used as the starting point in training. In the deep feature based method, the feature map of a certain layer of the pre-trained network was used as imaging variables (features) and a traditional classifier, such as support vector machine, was trained using the extracted features.

\section{Materials and Methods}

\subsection{Patient Population}

Following Institutional Review Board approval, we reviewed patient data at our institution from January 1, 2000 to March 23, 2014 to identify patients who underwent MRI and were diagnosed with breast cancer. We identified women satisfying the following criteria: CNB-confirmed DCIS without invasive cancer, a preoperative bilateral breast MRI with a nonfat-saturated T1-weighted MRI sequence, no neoadjuvant therapy before breast MRI, and no prior history of breast cancer. Using these criteria, we identified 131 patients in total, of whom 35 were upstaged to invasive cancer at surgery. Patient demographics for the cohort are illustrated in Table \ref{tab:characteristic}.
\begin{table}[]
\centering
\caption{Clinicopathologic characteristics of DCIS.}
\label{tab:characteristic}
\begin{tabular}{ll}
Characteristic    & No. of Patients (\%) \\ \hline
Age               &                      \\
\hspace{3mm}30-40             & 18(13.7)             \\
\hspace{3mm}40-50             & 42(32.1)             \\
\hspace{3mm}50-60             & 42(32.1)             \\
\hspace{3mm}60-70             & 22(16.8)             \\
\hspace{3mm}70-80             & 7(5.3)               \\
Race              &                      \\
\hspace{3mm}White             & 100(76.3)            \\
\hspace{3mm}Black             & 22(16.8)             \\
\hspace{3mm}Others            & 6(4.6)               \\
\hspace{3mm}N/A               & 3(2.3)               \\
Menopausal status &                      \\
\hspace{3mm}Pre               & 61(46.6)             \\
\hspace{3mm}Post              & 68(51.9)             \\
\hspace{3mm}N/A               & 2(1.5)               \\
ER status         &                      \\
\hspace{3mm}Neg               & 26(19.8)             \\
\hspace{3mm}Pos               & 99(75.6)             \\
\hspace{3mm}N/A               & 6(4.6)               \\ \hline
\end{tabular}
\end{table}

\subsection{MR Imaging}

All patients in this study underwent MR imaging of the breast, where preoperative breast axial dynamic contrast-enhanced MRI scans were acquired using 1.5T and 3.0T scanners in the prone position. For each patient, there was a pre-contrast sequence and at least 3 post-contrast sequences. The information about the scanners and the MRI protocols are listed in Table~\ref{tab:protocols}.
\begin{table}[]
\centering
\caption{Scanner information and MRI protocols.}
\label{tab:protocols}
\begin{tabular}{llllll}
Scanner         & Field Strength(T) & FOV(cm) & Matrix Size & TE (msec) & TR (msec) \\ \hline
Signa HDx       & 3.0               & 34      & 350*350     & 2.4       & 5.7       \\
Signa HDx/HDxt  & 1.5               & 38      & 350*350     & 2.4       & 5.3       \\
MAGNETOM Trio   & 3.0               & 36      & 448*448     & 1.4       & 4.1       \\
MAGNETON Avanto & 1.5               & 36      & 448*448     & 1.3       & 4.0       \\ \hline
\end{tabular}
\end{table}

\subsection{Image Annotation}
\begin{figure}
  \centering
  \includegraphics[width=.3\textwidth]{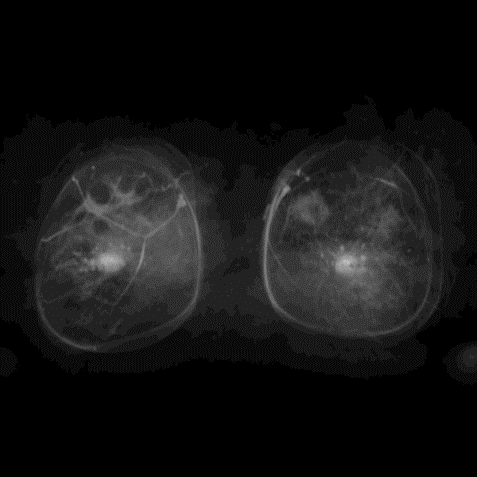}
  \includegraphics[width=.3\textwidth]{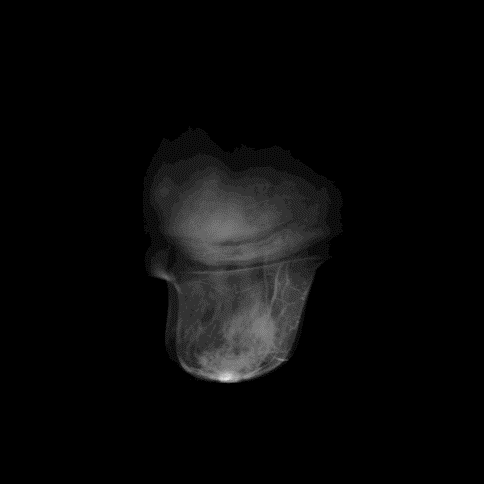}
  \includegraphics[width=.3\textwidth]{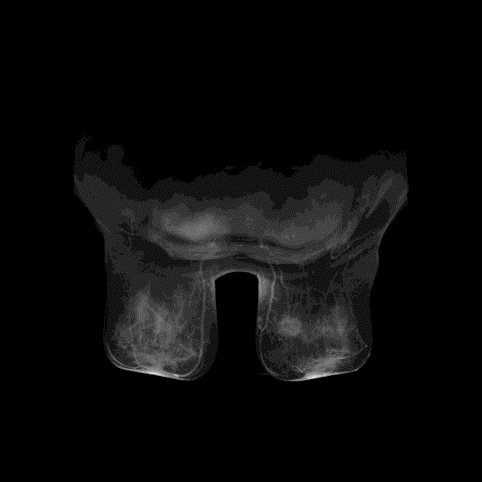}
  \\
  \caption{A visualization of the MRI with 3 orthorhombic views.}\label{fig:vis}
\end{figure}

One fellowship-trained breast-imaging radiologist with 1 year of experience at our institution annotated the images. We provided the reader with a T1-weighted pre-contrast fat-saturated sequence, T1-weighted post-contrast fat-saturated sequence, and the sequence showing the subtraction between these two sequences. Additional information was also provided to the reader for further assistance: breast CNB location of the DCIS collected from the medical record and access to the MRI exam in our institution's picture archiving and communication system (PACS). For each case, a compact 2D bounding box was drawn around the lesion region by the reader using a graphical user interface (GUI). After the reader reviewed the case, the range of slices that encompassed the totality of the lesion was chosen, resulting in the 3D bounding box for each tumor.  A visualization of the MRI with 3 orthorhombic views is shown in Figure~\ref{fig:vis}.

\subsection{Classification Framework}
Our approach was developed to classify each tumor as containing an invasive component or not, and thus the considered task is a binary classification task. The classification steps for a given image were as follows: (1) preprocessing of the image, (2) extracting patches to represent the tumor, (3) application of a deep learning-based classifier. The details of the steps are described below.

\subsection{MR Imaging Pre-processing and Tumor Patch Extraction}
Since multiple scanners were used to acquire the MRIs, there were several different spatial resolutions in our imaging data. Therefore, the first step was to register all the MRIs to achieve the same pixel spacing in x-y coordinates. The most common pixel spacing was chosen, and MRIs with other values of the pixel spacing parameter were scaled using bilinear interpolation.

\begin{figure}
  \centering
  \includegraphics[width=.95\textwidth]{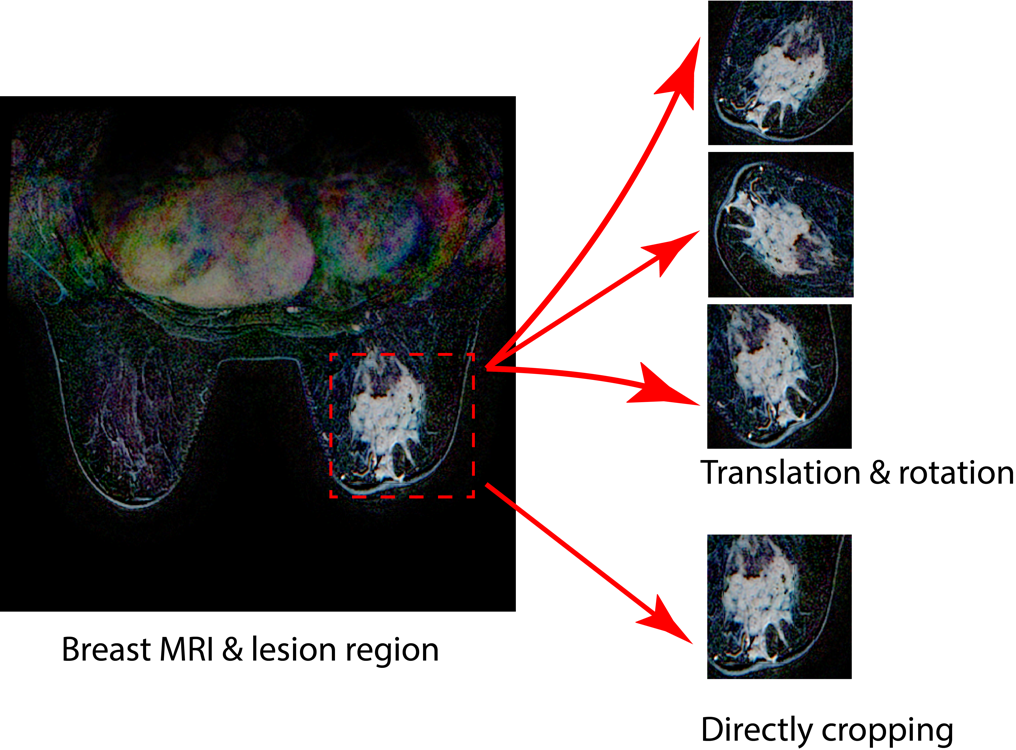}\\
  \caption{Patch extraction and data augmentation.}\label{fig:patch}
\end{figure}

For each patient, there were 4 sequences scanned at different times before and after the administration of contrast. We used the patch-based method that extracted patches from each slice as training samples. For each patient, the number of slices of the MRI was between 100 and 200, so there can be multiple samples for each patient. Since we used a network pre-trained on natural images that has three channels (red, green, and blue), we concatenated $S'-S_p$ ,$S''-S_p$ and $S'''-S_p$ as 3-channel image where $S'$, $S''$ and $S'''$ are 3 post-contrast sequences and $S_p$ is the pre-contrast sequence. For the same patient, the size of the lesion might differ in different slices. In order to eliminate the interference of non-lesion regions, only slices which contain lesion whose area of the bounding box is larger than 100 pixels were chosen for further processing. Then the center of the lesion was calculated from its bounding box. To further generate more samples, except the patch centered at the lesion center, 5 additional samples were generated by random rotations. We chose 120$\times$120 as the patch size according to the statistics of our DCIS dataset: the longest edge of the lesion's bounding box ranges between 40 and 160 for 80\% of the patients, so this size could cover most part of the lesion while the proportion of the lesion in the patch was not too small for some cases. The patch extraction and data augmentation strategy are illustrated in Figure~\ref{fig:patch}. Finally, there were 30426 patches generated in total, which is a reasonable number for transfer learning and deep features approach.

\subsection{Deep Learning Platform}
We used deep learning in two ways: transfer learning and off-the-shelf deep features. We chose GoogleNet7 which is one of the most representative network in image classification as the basis of our analysis. The details of the training are presented below. For our experiments, we used the CAFFE~\cite{Jia2014} framework on a desktop computer with a NVIDIA GTX 1080 GPU.

\subsection{Transfer Learning}
In our transfer learning approach, we chose GoogleNet pre-trained on ImageNet, which consists of convolution layers, Inception layers, and fully connected layers. Though convolution layers can deal with arbitrary input sizes, fully connected layers can only process fixed input dimension. To fit the input size of the GoogleNet and reduce the problem of overfitting in training, all patches extracted from the MRI images were resized to 256$\times$256$\times$3 using bilinear interpolation, and a 224$\times$224$\times$3 randomly sampled sub-image was cropped during each epoch in the training phase. The output dimension of the last fully connected layer of pre-trained GoogleNet is 1000 since there are 1000 categories in ImageNet. Here, we modified that dimension to 2 for our binary classification task. The last fully connected layer was completely re-initialized randomly while all the other layers maintained their weights from the pre-training. Since shallow layers correspond to more general but low-level image features while deeper layers are higher level but task-specific, the learning rate of deeper layers should be set larger than that of shallow layers. In our implementation, the learning rate for the final fully connected layer was set to 0.001 and the learning rate of all the other layers was set to 0.0001. Following a seminal publication in the field6 we used the following parameters in training: the maximum number of iterations was set to 200000, gamma was set to 0.96, momentum was set to 0.9, batch size was set to 32. We used stochastic gradient descent (SGD) for training.

\subsection{Off-the-shelf Deep Features}
Since our task is to classify the upstaging of the DCIS, we chose ImageNet pre-trained networks using a classification task, as similar task should share more common features. More specifically, we chose the GoogleNet pre-trained on ImageNet and used the feature map of the last fully connected layer~\cite{Zhou2014} as the primary feature map and trained an SVM on those features. As each individual patch can generate a unique feature vector, there were tens of thousands training samples for SVM. Since kernel functions have great influence on the performance of the SVM, several different kernel functions were evaluated during our experiments. The feature maps of different layers were also evaluated for comparison.

\subsection{Model Evaluation}
We used 10-fold cross-validation scheme to evaluate the trained classifiers. The data was split by patients, ensuring that data (multiple patches) for one patient were fully contained in the training set or in the test set (never in both). Note that there are multiple slices for each patient. We picked 5 slices that have the 5 largest lesion areas, and for each slice the patch centered at the lesion's bounding box center without translation and rotation was chosen for the test. Then, scores of the 5 patches were averaged. The AUCs in the 10-folds were averaged to arrive at the final performance evaluation. Each fold had a similar rate between positive (upstaged) samples and negative (not upstaged) samples. Confidence intervals for AUCs were estimated using the bootstrapping strategy.

\section{Results}
\label{gen_inst}

\begin{figure}
  \centering
  \includegraphics[width=.55\textwidth]{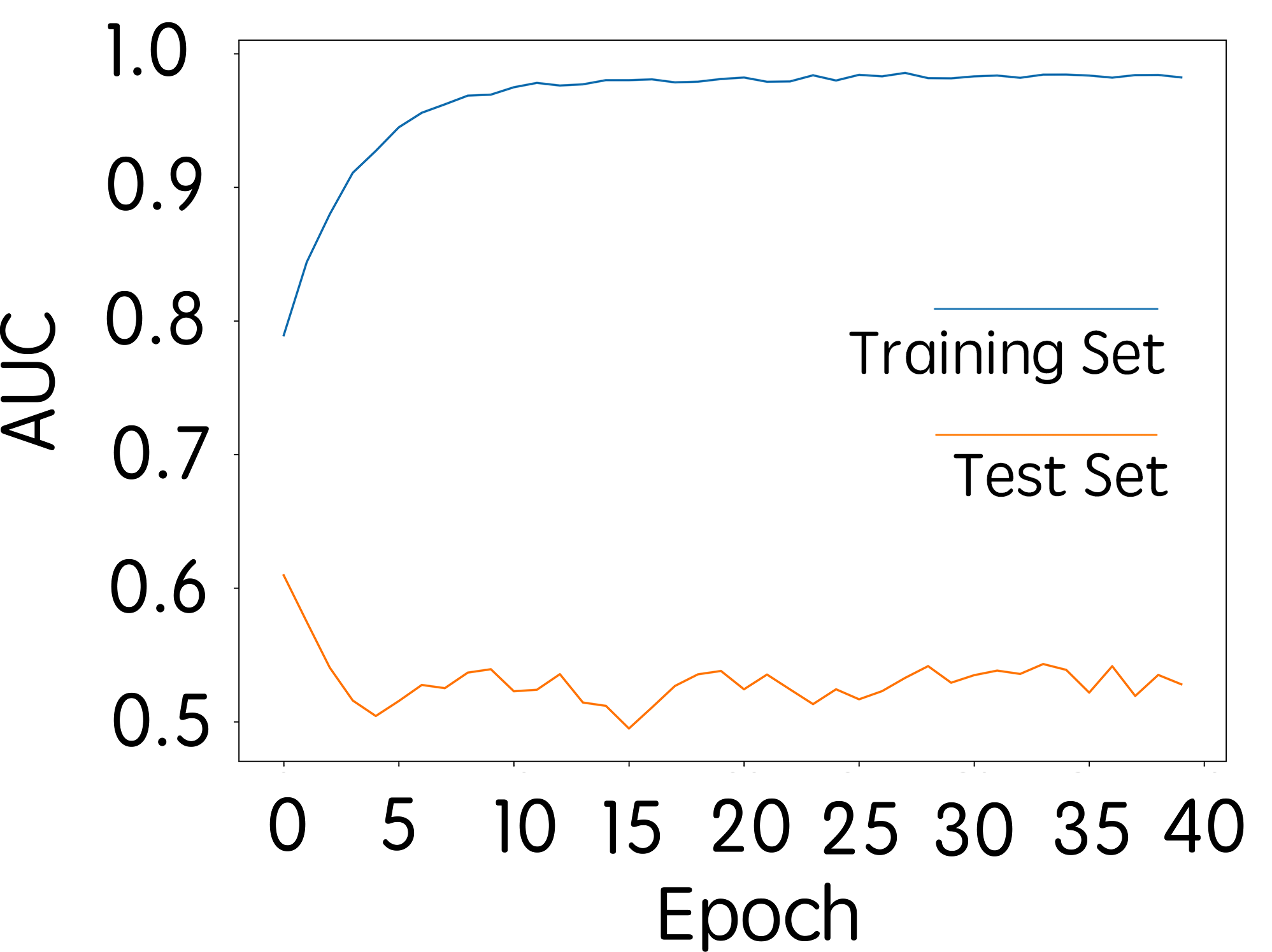}\\
  \caption{AUC training curves for transfer learning. Orange line is the AUC for the test set while the blue line is the AUC for the training set.}\label{fig:trans}
\end{figure}

The transfer learning approach resulted in a final AUC of 0.53 (95\% CI: 0.41-0.62). The AUC-based training curve for the network is shown in Figure~\ref{fig:trans}. For the performance on the training set, the AUC increased during the first 10 epochs, and reached the stable value of 0.96 after 10 epochs. On the contrary, the performance on the test set varied around 0.53, showing performance close to chance. This indicates a strong overtraining effect when transfer learning is applied to our problem.

For the off-the-shelf deep feature based approach, the highest AUC obtained was 0.70 (95\% CI: 0.58-0.79) with an SVM using polynomial kernel. To further verify this result, we altered the patch size in the patch extraction step and explored different kernel functions in SVM. Representative results are illustrated in Table~\ref{tab:patch_size}. To achieve the best suitable patch size, we fixed the kernel function in SVM to polynomial, and found that 80 was the best patch size for our dataset. The result of using 80 as the patch size and polynomial kernel SVM is illustrated in row 4 of Table~\ref{tab:patch_size}. Results with different patch sizes but the same kernel function in SVM are illustrated in row 2, row 6, and row 7 of Table~\ref{tab:patch_size}. To choose the best kernel function in SVM, we fixed the patch size and evaluated the performance of different kernel functions in SVM. The results using the same patch size with different kernel functions in SVM are illustrated in row 3, row 4, and row 5 of Table~\ref{tab:patch_size}. These results indicate that polynomial kernel function has the highest performance for our DCIS dataset.

\begin{table}[]
\centering
\caption{AUCs with different patch sizes and kernels of SVM.}
\label{tab:patch_size}
\begin{tabular}{llll}
Patch Size & Kernel Function & Training AUC & Test AUC \\ \hline
75         & poly            & 0.81         & 0.68     \\
80         & rbf             & 0.84         & 0.55     \\
80         & poly            & 0.86         & 0.70     \\
80         & linear          & 0.77         & 0.63     \\
85         & poly            & 0.85         & 0.67     \\
100        & poly            & 0.86         & 0.65     \\ \hline
\end{tabular}
\end{table}

\begin{figure}
  \centering
  \includegraphics[width=.95\textwidth]{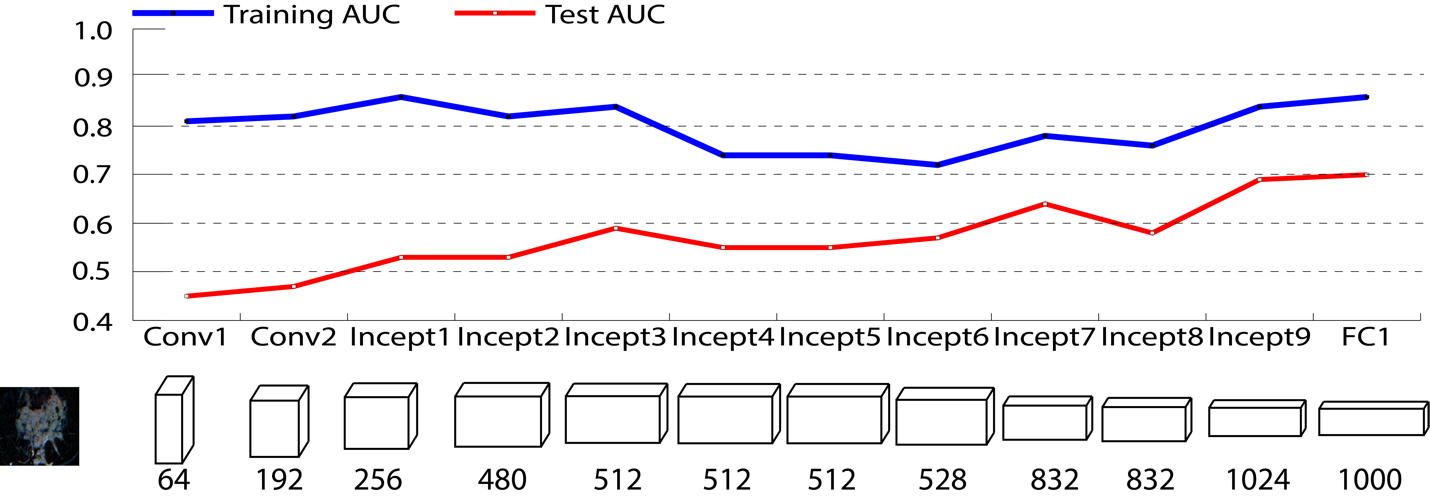}\\
  \caption{AUCs for different feature maps.}\label{fig:feature_map}
\end{figure}

Our primary results were obtained by using the feature map of the last fully connected layer. Additionally, we investigated the performances of different layers' feature maps of GoogleNet. Specifically, the GoogleNet consists of 2 convolution layers, 9 Inception layers and 1 fully connected layer. Except the last fully connected layer, feature maps of other layers are 2-dimensional matrices in each feature channel, making the overall dimension too large. Thus, for those feature maps, we used a max pooling strategy in the image plane, so the dimension of the output feature map equals to the number of original feature channels. For example, the output of the first convolution layer is a 64$\times$112$\times$112 matrix, and the max pooled feature vector is a 64-dimensional vector. For each layer's max pooled feature map, an SVM with polynomial kernel was trained as a classifier. The results using different layers' feature maps are illustrated in Table~\ref{tab:layers} and Figure~\ref{fig:feature_map} respectively. From these results we can conclude that the last fully connected layer has the highest performance in our task.

\begin{table}[]
\centering
\caption{Performance of feature maps from different layers.}
\label{tab:layers}
\begin{tabular}{lllllllllllll}
                                                         & Conv1 & Conv2 & Incp1 & Incp2 & Incp3 & Incp4 & Incp5 & Incp6 & Incp7 & Incp8 & Incp9 & FC1  \\ \hline
\begin{tabular}[c]{@{}l@{}}Feature\\ Length\end{tabular} & 64    & 192   & 256    & 480    & 512    & 512    & 512    & 528    & 832    & 832    & 1024   & 1000 \\
\begin{tabular}[c]{@{}l@{}}Training\\ AUC\end{tabular}   & 0.81  & 0.82  & 0.86   & 0.82   & 0.84   & 0.74   & 0.74   & 0.72   & 0.78   & 0.76   & 0.84   & 0.86 \\
\begin{tabular}[c]{@{}l@{}}Test\\ AUC\end{tabular}       & 0.45  & 0.47  & 0.53   & 0.53   & 0.59   & 0.55   & 0.55   & 0.57   & 0.64   & 0.58   & 0.69   & 0.70 \\ \hline
\end{tabular}
\end{table}

\begin{figure}
  \centering
  \includegraphics[width=.55\textwidth]{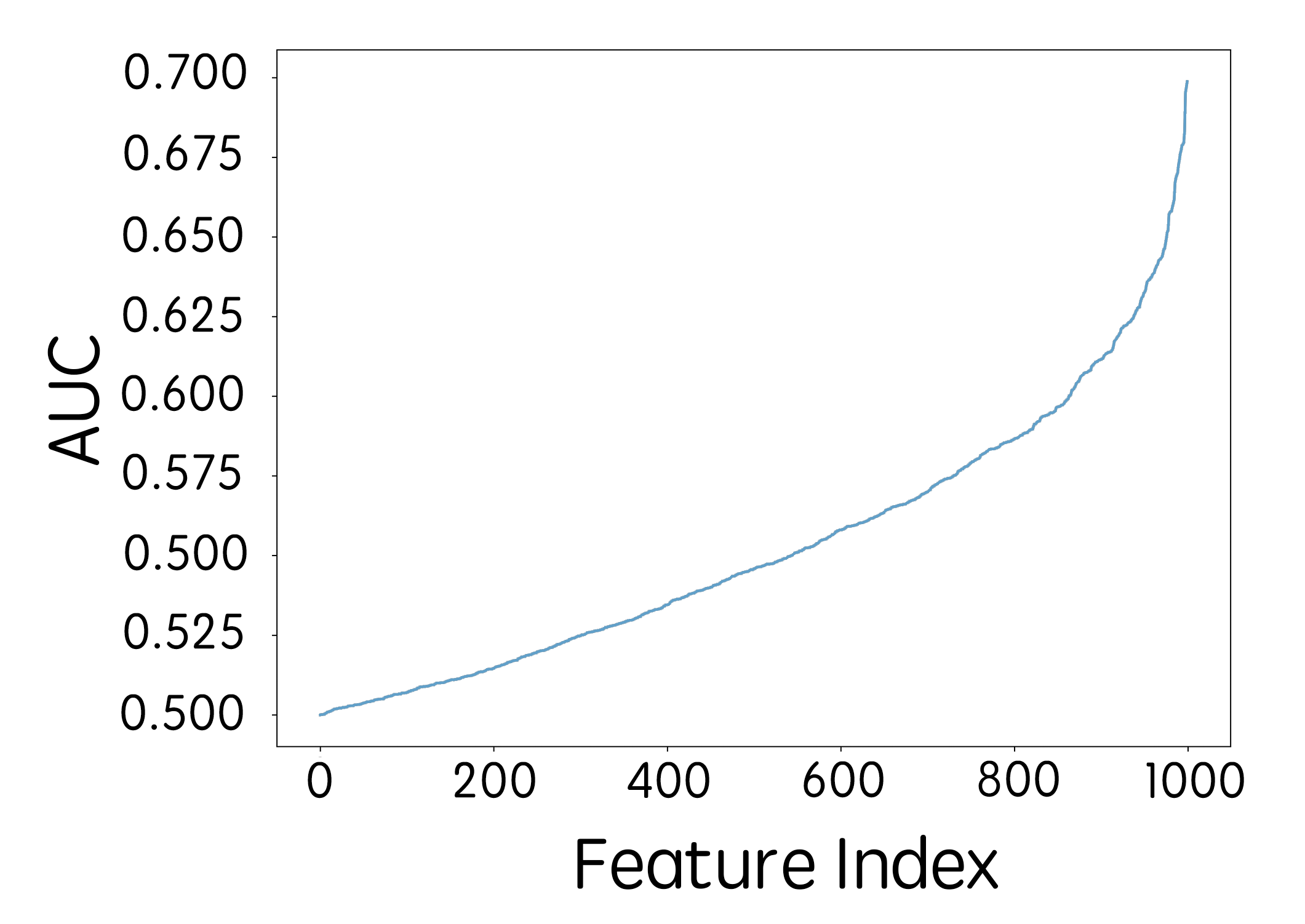}\\
  \caption{AUCs for each individual feature. Features have been sorted by AUC.}\label{fig:indiv}
\end{figure}

Finally, we also evaluated the classification performance in terms of AUC for each individual feature from the set of 1000 deep features used in our primary multivariate (SVM) model. In order to perform the classification, the value of each individual feature was regarded as the predicted score. The AUCs for all the features, ordered by the performance, are illustrated in Figure~\ref{fig:indiv}. The AUC of each individual feature ranged from 0.5 to 0.69. This indicates that the best performance, which was obtained by combining all these features, is slightly higher than the best performance by using single feature only.

\section{Discussion}
\label{headings}

Prediction of occult invasive disease in DCIS is of high importance in the current clinical practice of breast cancer treatment. With accurate prediction of upstaging to invasive using MRI image features before surgery, DCIS patients would be given more personalized and appropriate treatment earlier. Furthermore, identifying invasive disease among DCIS patients is crucial when selecting patients for active surveillance. Since in this new treatment paradigm the patient does not undergo surgery, missing an invasive component could have significant negative consequences to the patient.

We studied the problem of predicting upstaging of DCIS using two deep learning methods: transfer learning and off-the-shelf deep features. The transfer learning approach suffered from overfitting and resulted in poor classification performance, while the off-the-shelf deep features approach achieved relatively good results. Results of the latter approach show promise of using deep learning approach to predict the upstaging of DCIS.

Our experiment with transfer learning indicated that insufficient training data can cause the overfitting problem. Although some previous studies have shown that pre-trained deep networks can generalize well to some other tasks~\cite{Zhou2014}, in our evaluation of transfer learning, the performance on training set reached near perfect while on the test set the performance resembled random guess. This implies that fine-tuning a pre-trained network on a new task still requires significant amount of data specific to the solved task which was not available in this study. Compared with the random weights initialization in training from scratch, transfer learning starts with better initial weights. Nevertheless sufficient data is also needed to train the weights in each layer of the network.

Two main factors contributed the promising results using deep features approach. First, the pre-trained networks can extract rich image features.
Specifically, since we used the pre-trained network on classification task, the extracted features were more suited for DCIS upstaging prediction. Second, the extracted features were used to train a traditional classifier (SVM) which is less prone to overfitting as compared with deep neural networks.
Using medical imaging for prediction of occult invasive disease in DCIS is a relatively new topic and limited literature is available. The gold standard is MRI review by highly trained breast imagers~\cite{WisnerDJ;HwangES;ChangCB;TsoHH;JoeBN;LessingJN;LuY;Hylton2013}, but there has been interest in exploring a more reproducible and automated process for this task. Recently, Harowicz et al.~\cite{Harowicz2017} showed that an individual hand crafted feature of MRI may be predictive of occult invasive disease. However no multivariate model was presented in this study. Shi et al.~\cite{Shi2017} showed that mammographic features could also potentially be used to distinguish DCIS from invasive cancer. Our study builds on these results by showing that deep learning methods which do not require design and selection of specific features could be used for this task despite a small sample size.

Our work had limitations. First, the evaluation result using 10-fold cross-validation is a common choice, however, the lack of an independent test set is a limitation. Second, the deep features were used as a "black box" which means that those features might not have a specific meaning that would be understandable to a clinician. Finally, the size of the analyzed dataset was limited. Future studies with a larger dataset are required to validate the results presented in our paper. Further, we acknowledge that an AUC of 0.70, while promising, is not sufficiently accurate for clinical use in separation from other currently used tools. Thus, further steps to optimize the performance of the model are planned. The primary step to improving performance of this classifier is a collection of a large multi-institutional dataset for training. This will allow for more optimal utilization of the power of convolutional neural networks. Furthermore, an improved normalization of the images from different imaging protocols might result in improvements in performance. Finally, while the imaging-based mode alone might not achieve a sufficiently high predictive performance, a combination of the imaging-based model with a pathology-based model could result in satisfactory prediction accuracy. In the long term, in order to incorporate these types of tools to the clinical setting, we must perform a validation of the optimized tool in a multi-institutional setting as well as place it in the proper context of other clinicopathologic variables.

In conclusion, we demonstrated that a deep learning-based model based on MR imaging showed predictive power for identifying occult invasive cancer in patients diagnosed with ductal carcinoma in situ using core needle biopsy. Further refinement of this has the potential to be of clinical value for patients with DCIS making treatment decisions for their disease.

\bibliography{iclr2017_conference}
\bibliographystyle{iclr2017_conference}

\end{document}